\documentclass{article}
\usepackage{spconf,amsmath,graphicx}
\usepackage{multirow}
\usepackage[font=small]{caption}
\usepackage{subcaption}
\usepackage{tikz}
\usepackage{amssymb}
\usepackage{amsmath}
\usepackage{etoolbox}
\usetikzlibrary{arrows}

\BeforeBeginEnvironment{tikzpicture}{\vskip-.3ex}
\AfterEndEnvironment{tikzpicture}{\vskip-1.7ex}

\newcommand\copyrighttext{\footnotesize \textcopyright 2018 IEEE. Published in the IEEE 2018 International Conference on Image Processing (ICIP 2018), scheduled for 7-10 October 2018 in Athens, Greece. Personal use of this material is permitted. However, permission to reprint/republish this material for advertising or promotional purposes or for creating new collective works for resale or redistribution to servers or lists, or to reuse any copyrighted component of this work in other works, must be obtained from the IEEE. Contact: Manager, Copyrights and Permissions / IEEE Service Center / 445 Hoes Lane / P.O. Box 1331 / Piscataway, NJ 08855-1331, USA. Telephone: + Intl. 908-562-3966.}
\renewcommand\copyrightnotice{%
\begin{tikzpicture}[remember picture,overlay]
\node[anchor=south,yshift=10pt] at (current page.south) {\fbox{\parbox{\dimexpr\textwidth-\fboxsep-\fboxrule\relax}{\copyrighttext}}};
\end{tikzpicture}%
}

\title{AESTHETICS ASSESSMENT OF IMAGES CONTAINING FACES}
\name{Simone Bianco, Luigi Celona, Raimondo Schettini}
\address{Department of Informatics, Systems and Communication\\
University of Milano-Bicocca\\
viale Sarca, 336, 20126, Milano, Italy}
\begin{document}
%
\maketitle
\copyrightnotice
\begin{abstract}
Recent research has widely explored the problem of aesthetics assessment of images with generic content. However, few approaches have been specifically designed to predict the aesthetic quality of images containing human faces, which make up a massive portion of photos in the web. This paper introduces a method for aesthetic quality assessment of images with faces. We exploit three different Convolutional Neural Networks to encode information regarding perceptual quality, global image aesthetics, and facial attributes; then, a model is trained to combine these features to explicitly predict the aesthetics of images containing faces. Experimental results show that our approach outperforms existing methods for both binary, i.e. low/high, and continuous aesthetic score prediction on four different databases in the state-of-the-art.
\end{abstract}
\begin{keywords}
Image aesthetics, Faces, Convolutional neural networks, Genetic algorithms
\end{keywords}
\section{Introduction}
Automatic image aesthetic assessment is a challenging task due to its fuzzy definition and its highly subjective nature. It represents an important criterion for visual content curation and it is useful in many applications such as image retrieval \cite{li2010towards,vonikakis2017probabilistic}, photo enhancement \cite{bhattacharya2010framework}, and image cropping \cite{ciocca2007self-adaptive-image,bianco2015-user-preferences,jin2016image}. Aesthetic assessment of images with generic content has been addressed in \cite{jin2016image,Bianco2016predicting-image,kao2017deep}. However, psychology research \cite{freeman2007photographer} showed that certain kinds of content are more attractive than others. In fact, professional photographers adopt different photographic techniques and have various aesthetic criteria in mind when taking different types of photos. Therefore, it is reasonable to design  features specialized in modeling aesthetic quality for different kinds of photos (e.g. \cite{luo2011content}).

\begin{figure}[t]
\centering
\resizebox{\columnwidth}{!}{
\begin{tabular}{cc}
\includegraphics{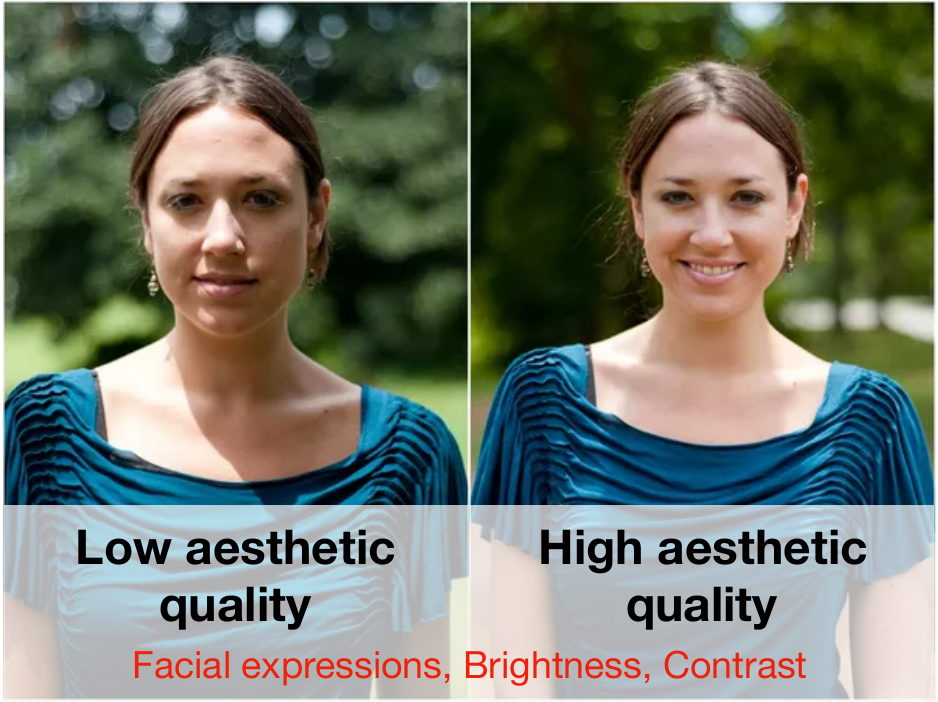}	\hspace{1em}
\includegraphics{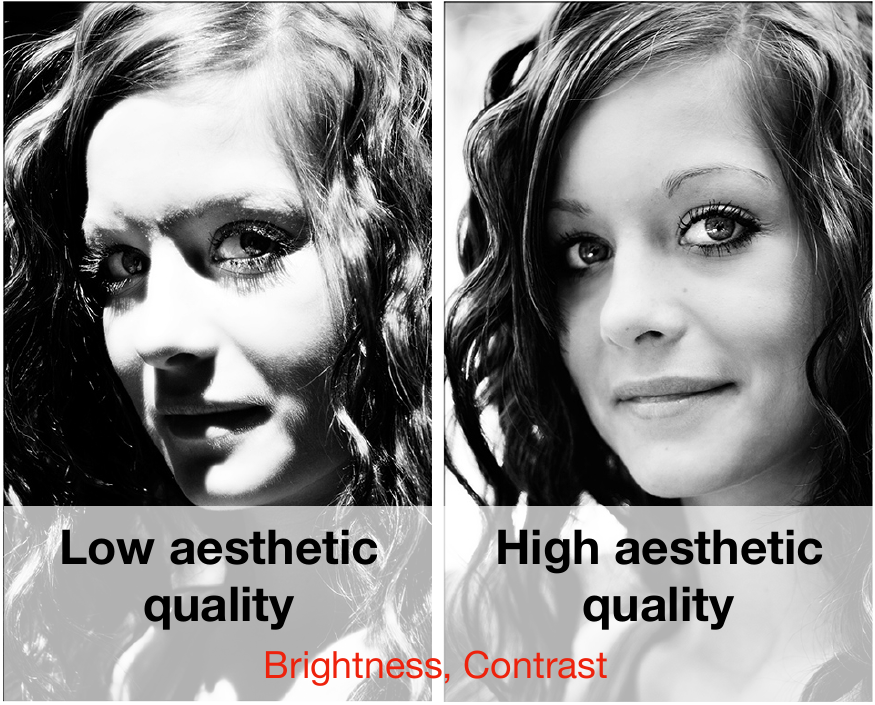}
\end{tabular}}
\caption{Face aesthetics represents the attractiveness of the photo shot. This takes into account aspects such as: facial expressions, brightness, contrast, etc.}
\label{fia-sample}
\end{figure}

In this paper we focus on aesthetic assessment of images containing human faces. The reasons are twofold: this category of photos makes up an important part of images on social media sites and media content repositories \cite{bianco2014adaptive,bakhshi2014faces}, and we have observed that the performance of generic content aesthetic assessment methods \cite{Bianco2016predicting-image} drop considerably when dealing with this type of images. It should be clear that although facial beauty and face aesthetics are two related concepts, the first reflects the attractiveness of the subject's face, while the second represents the attractiveness of the photo containing the subject's face (see for example Fig. \ref{fia-sample}). 

Li \textit{et al.} \cite{li2010aesthetic} evaluated the performance of several categories of features related to aesthetics such as pose, face locations and photo composition on their own dataset of photos with faces. Males \textit{et al.} \cite{males2013aesthetic} exploited a support vector machine for aesthetic quality categorization trained on the combination of global (e.g. contrast and hue distribution of the whole image) and local features (e.g. sharpness and blown-out highlights only of facial region). Their experiments have been carried out on a set of photo collected from Flickr and manually labeled by five people as being aesthetically appealing or not. Lienhard \textit{et al.} 
\cite{lienhard2014photo,lienhard2015predict} proposed a new database, called Human Faces Score (HFS), and developed a method based on the selection of low-level features extracted from several regions for both aesthetic quality categorization of portrait images (i.e. low or high) and continuous aesthetic score prediction. Recently, in \cite{kairanbay2016aesthetic} a compositional-based augmentation scheme has been used to train a deep convolutional neural network (DCNN) on a portrait subset of the AVA dataset for binary aesthetic classification.
\section{Facial image aesthetic estimation}
\label{method}
\begin{figure*}
\centering
\includegraphics[width=.95\textwidth]{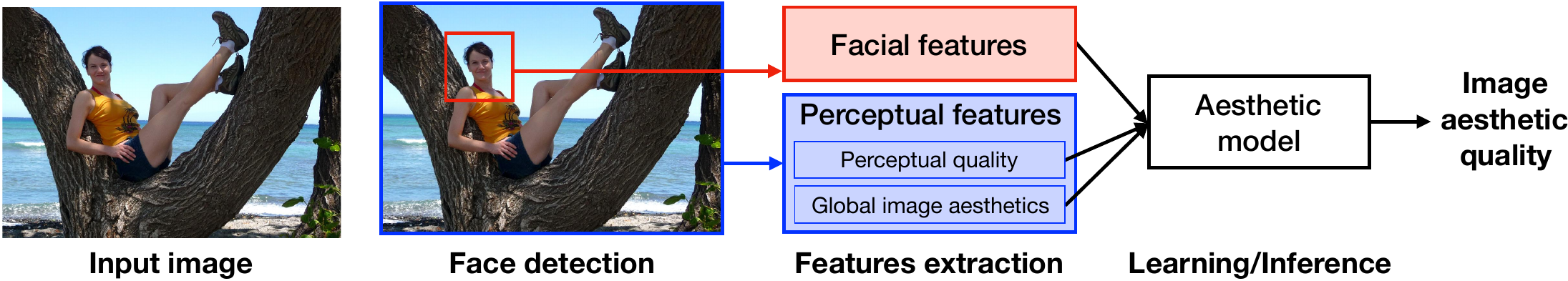}
\caption{Overview of the proposed method.}
\label{fia-pipeline}
\end{figure*}
In this section we describe the proposed method for aesthetic quality assessment of images with faces. The proposed method is depicted in Fig. \ref{fia-pipeline}: given a photo, first the largest face is detected, then features are extracted from the whole image and the face region, and finally the trained model is applied for aesthetic quality estimation of the photo.
\subsection{Face detection}
\emph{DLib}'s face detector \cite{dlib09} is used to localize the face region. The size of the detected bounding box is then increased of 10\% in order to include also a portion of the shoulders.
\subsection{Features extraction}
Aesthetic quality of photos with generic content as well as the aesthetics of photos with faces depend upon several perceptual properties. Furthermore, face attributes provide fundamental information for the aesthetic evaluation of this specific category of photos. In this paper, we use state-of-the-art CNNs for encoding both perceptual image-related and face properties.

\textbf{Perceptual features.} As highlighted in many previous works, aesthetic quality is strongly influenced by several dimensions such as composition, colorfulness, spatial organization, emphasis, and depth. We consider two pre-trained CNNs for image quality assessment and generic content aesthetics assessment, proposed in authors' previous works, in order to encode such information about the whole image (face and background). Specifically, the DeepBIQ model \cite{bianco2016onthe} (shortly IQ), that is a CNN model trained for blind image quality assessment, is considered for encoding \textit{perceptual quality} metrics such as noise, exposure, quality, JPEG quality, and sharpness. While, the DeepIA model \cite{Bianco2016predicting-image} (shortly IA), which is a CNN trained for generic content aesthetic assessment, is used to extract features related to \textit{global image aesthetics} concepts, such as brightness, contrast, color, etc.

Both IA and IQ are 4,096-dimensional feature vectors obtained by considering the activation of the last fully-connected layer immediately before the regression layer.

\textbf{Facial features.} In photos containing faces, observers mainly focus on face regions. Intuitively, face attributes such as facial expressions, the presence of makeup or the presence of accessories are closely related to the aesthetics of this specific category of photos. Therefore, we consider a set of features able to accurately describe the face. To this aim, we use the Alignment-Free Facial Attribute Classification Technique (AFFACT) \cite{guenther2017affact}, shortly FA, a CNN model trained for the estimation of 40 facial attributes. The 2,048-dimensional vector corresponding to the activations of the fully-connected layer before the classification layer are used as features.
\subsection{Features fusion and learning procedure}
Previously extracted features are fused and then exploited for the learning procedure following two different strategies.

The first includes linear concatenation as fusion technique, followed by a linear support vector machine (SVM) trained to estimate the portrait aesthetic quality. Since the resulting feature vectors have a huge number of features ($10,240$ when all the features are concatenated), some of which might be redundant, the second strategy proposed also includes a feature selection step. Feature selection refers to the task of identifying relevant features useful to fit accurate models. In this work, we propose a method based on genetic algorithms (GA) to jointly identify a subset of features from the whole feature vector and to optimize a prediction model. The GA is build to solve a mixed integer problem where some variables are restricted to take only integer values. Real-valued variables are the weights of the linear model which maps features to an aesthetic prediction, while the boolean-valued variables discriminate relevant features from the non-relevant ones. A chromosome is then represented as $(i_0 i_j ... i_{N_f},r_0 r_j ... r_{N_f},b)$, where $i_j \Rightarrow \{x \in \mathbb{Z} : 0 \le x \le 1 \}$ are binary values coordinating features selection, $r_j \in \mathbb{R}$ are the weights, $b \in \mathbb{R}$ is the bias, $x_j$ are the features, $j \in [0, N_f]$, and $N_f$ is the total number of features. Aesthetic quality is predicted through the following equation:
\begin{align}
pred = \sum_{j=0}^{N_f}{x_j(i_jr_j) + b}.
\label{pred_clas}
\end{align}
The fitness function used for classification tries to minimize the Hinge loss, while the fitness function for regression is the $\text{Smooth-L}_{1}$ loss (defined in \cite{girshick2015fast}).
\section{Experiments}
\label{experiments}
In this section, the evaluation procedure, the considered databases, the experiments and the results are detailed.
\subsection{Performance evaluation}
For the experiments the same evaluation procedure adopted in \cite{lienhard2015predict} is followed. More in detail, for each experiment 10-fold cross validation is performed by randomly selecting the training and testing images. This procedure is repeated 10 times to avoid sampling bias.

Classification performance is evaluated in terms of Good Classification Rate (\textit{GCR}) that is defined as the ratio between the number of images correctly classified and the number of test images. This is equal to compute classification accuracy.

Regression performance is evaluated in terms of Pearson's Linear Correlation Coefficient (\textit{LCC}) between the predicted and the ground-truth aesthetic scores. The average of both GCR and LCC  across the 10 rounds is reported.
\subsection{Portrait images databases}
In this section the publicly available databases for aesthetic assessment of images with faces are described. Databases consist of images containing people or groups of people gathered from online photo databases or photo sharing websites (e.g. Flickr, DPChallenge). Given that these photos are collected in real scenarios they present a wide range of subjects, facial appearance, illumination and imaging conditions.

\textbf{CUHKPQ.} The CUHKPQ \cite{tang2013content} is a database manually annotated for image aesthetics categorization (respectively high and low). It consists of 17,673 images organized in seven different categories. In this work, only images belonging to the ``human'' category are considered. There are 3,148 photos and some sample images are shown in Figure \ref{cuhkpq-samples}.

\textbf{Human Faces Scores (HFS).} The Human Faces Scores (HFS) \cite{lienhard2014photo} database contains 250 headshot photos. Specifically, 7 images of 20 different people, and 110 additional portrait images have been collected. Face images of one subject are given in Figure \ref{hfs-samples}. Each image has been rated by 25 human observers on a scale with values ranging between 1 and 6 (the highest aesthetic quality).

\textbf{Face Aesthetics Visual Analysis.} The Face Aesthetics Visual Analysis (FAVA) database is a subset of the large-scale AVA dataset \cite{murray2012ava} containing various images with faces. Each picture is associated with a value between 1 and 10 (the highest quality) corresponding to the average of around 210 collected individual scores. Samples are shown in Figure \ref{fava-samples}.

\textbf{Flickr database.} The Flickr database has been gathered on Flickr for general aesthetic assessment \cite{li2010towards}. It consists of 500 images associated to a ground-truth score between 0 and 10, where 10 means high quality. Photos are either portraits or group of faces. According to \cite{lienhard2015predict} only the biggest detected face is considered in each picture. Figure \ref{flickr-samples} shows samples from the database.
\begin{figure}[t]
\begin{subfigure}{\columnwidth}
\centering
\includegraphics[height=50px]{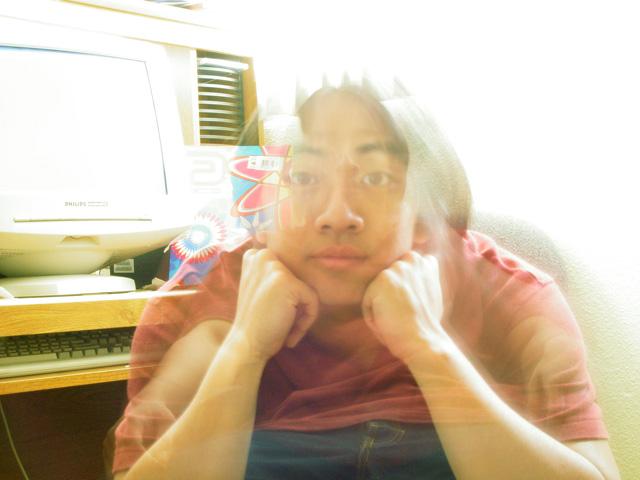}	
\includegraphics[height=50px]{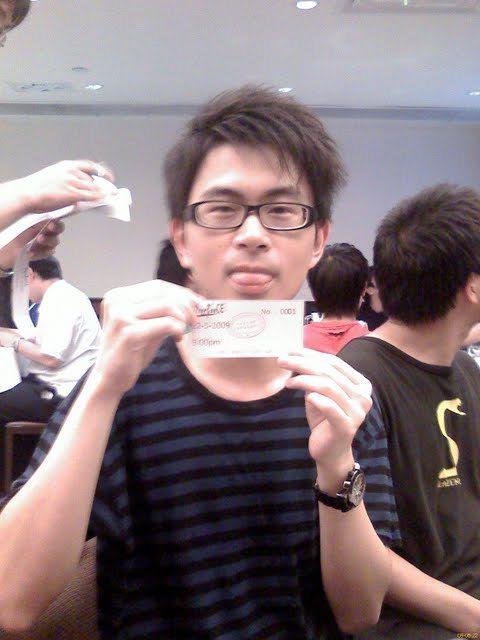}	
\hspace{1.2em}
\includegraphics[height=50px]{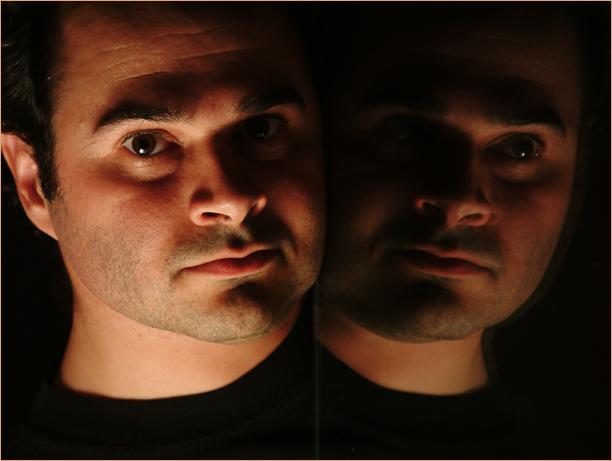}	
\includegraphics[height=50px]{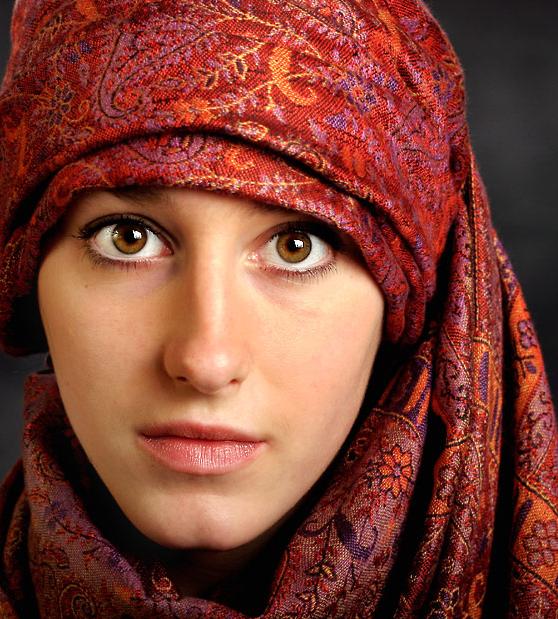}	
\begin{tikzpicture}
\node[draw=none,fill=none] (B) at (210:2.6) {low};
\node[draw=none,fill=none] (C) at (330:2.6) {high};
\end{tikzpicture}
\caption{Face images from the CUHKPQ database.}
\label{cuhkpq-samples}
\vspace*{1.2em}
\end{subfigure}
\begin{subfigure}{\columnwidth}
\centering
\includegraphics[height=50px]{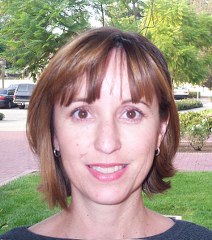}	
\includegraphics[height=50px]{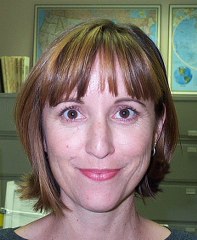}	
\includegraphics[height=50px]{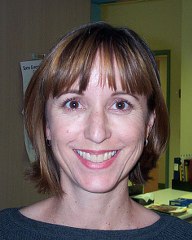}	
\includegraphics[height=50px]{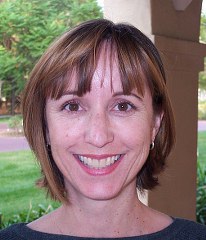}	
\includegraphics[height=50px]{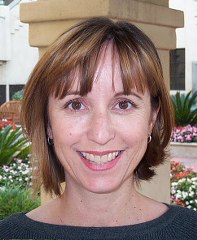}	
\begin{tikzpicture}
\node[draw=none,fill=none] (B) at (210:4.6) {1};
\node[draw=none,fill=none] (C) at (330:4.6) {6};
\draw[latex'-latex',double] (B) -- (C);
\end{tikzpicture}
\caption{Face images of one subject from the HFS database.}
\label{hfs-samples}
\vspace*{1.2em}
\end{subfigure}
\begin{subfigure}{\columnwidth}
\centering
\includegraphics[height=50px]{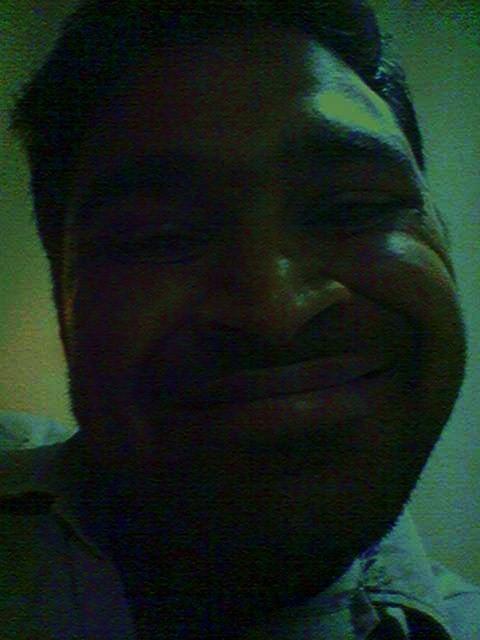} 
\includegraphics[height=50px]{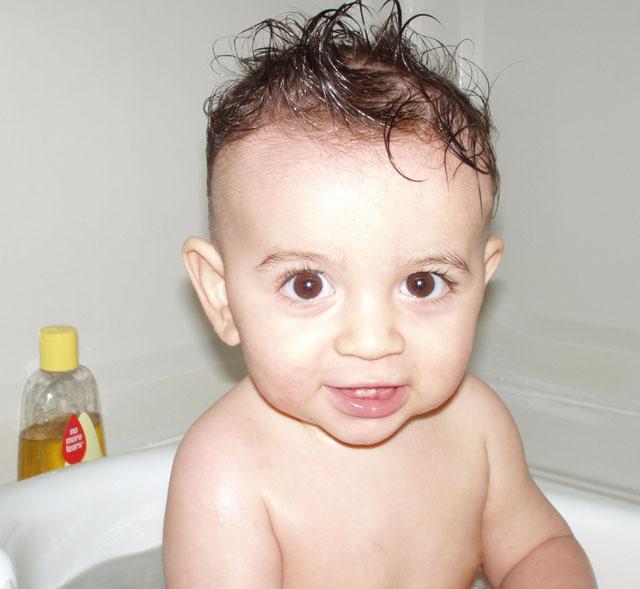}	
\includegraphics[height=50px]{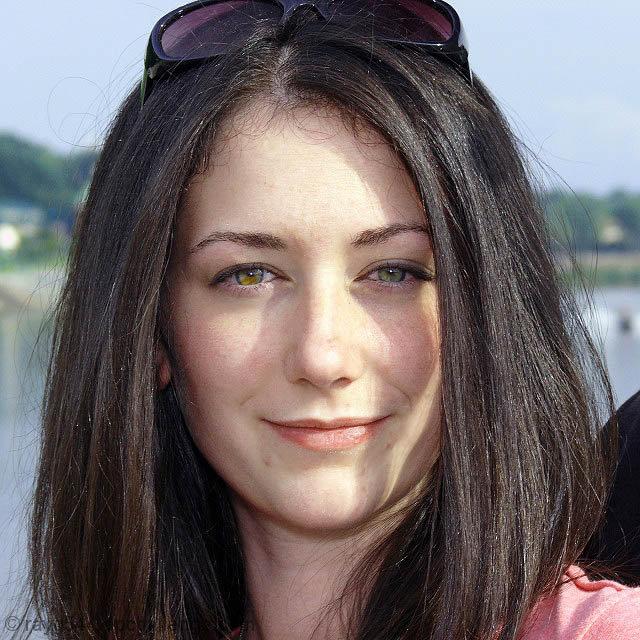}	
\includegraphics[height=50px]{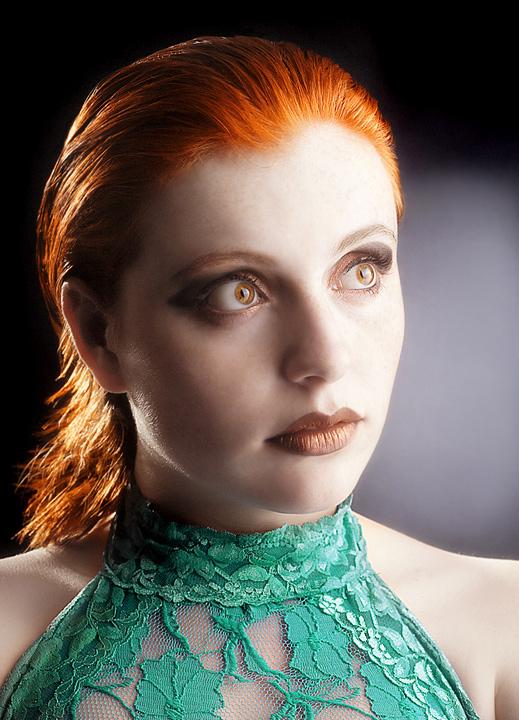}	
\includegraphics[height=50px]{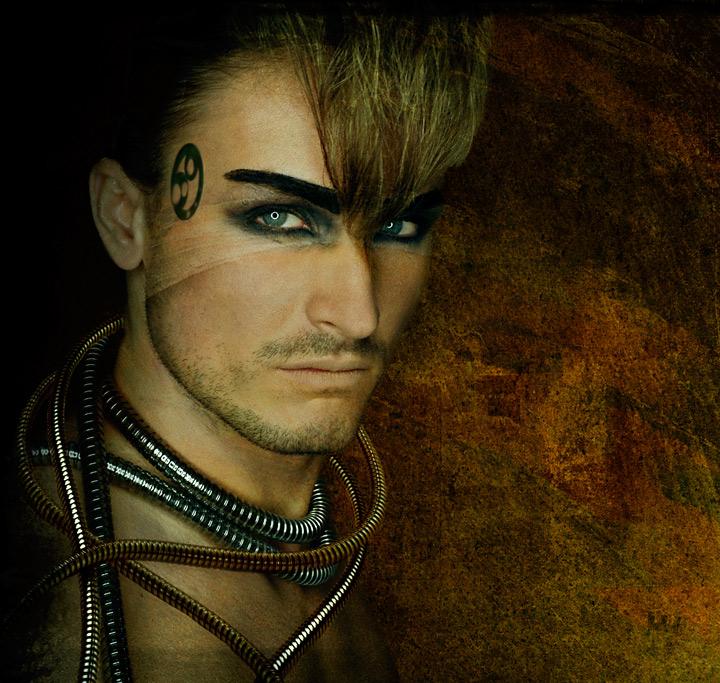}	
\begin{tikzpicture}
\node[draw=none,fill=none] (B) at (210:4.6) {1};
\node[draw=none,fill=none] (C) at (330:4.6) {10};
\draw[latex'-latex',double] (B) -- (C);
\end{tikzpicture}
\caption{Face images from the FAVA database.}
\label{fava-samples}
\vspace*{1.2em}
\end{subfigure}
\begin{subfigure}{\columnwidth}
\centering
\includegraphics[height=50px]{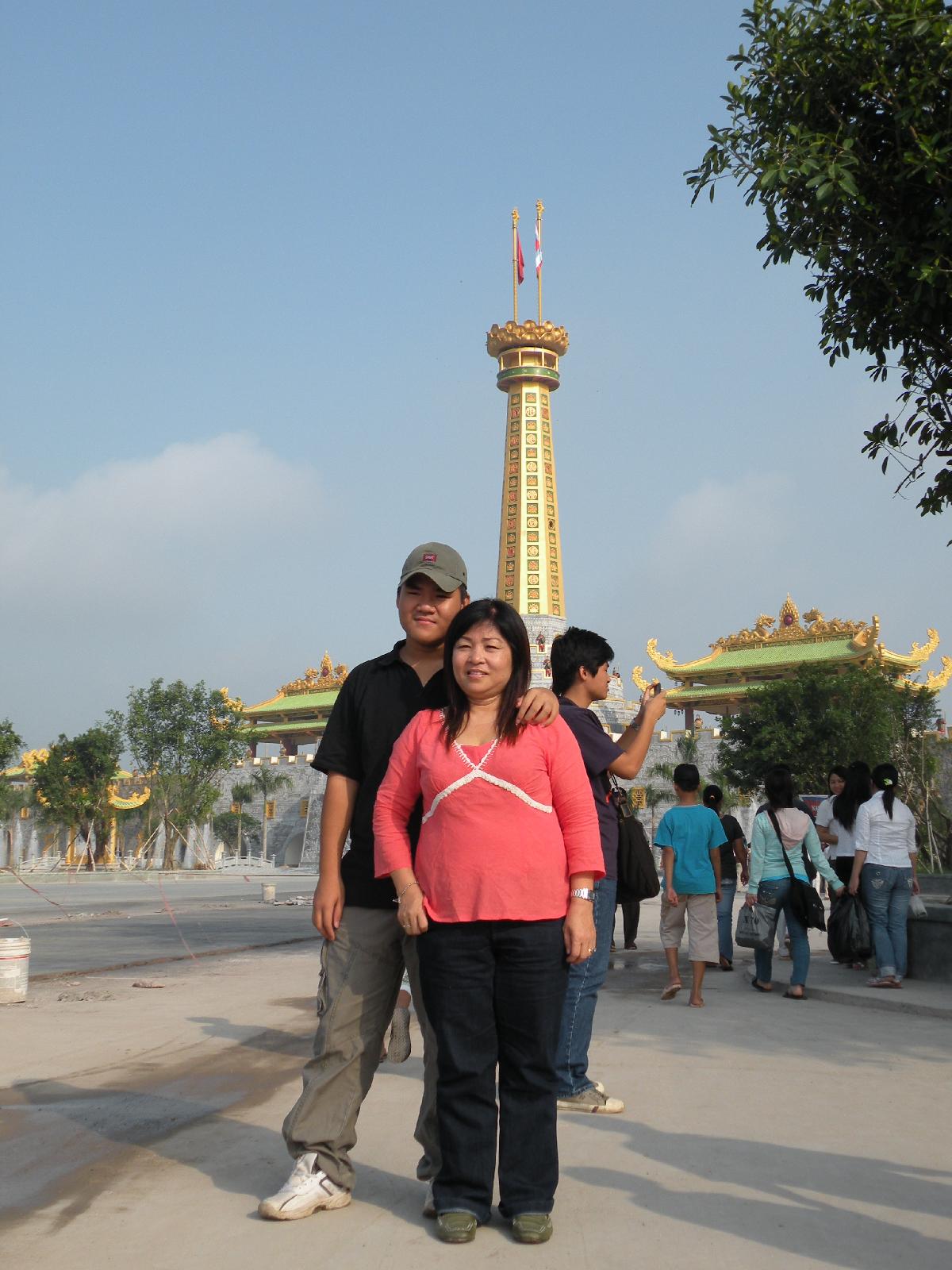}	
\includegraphics[height=50px]{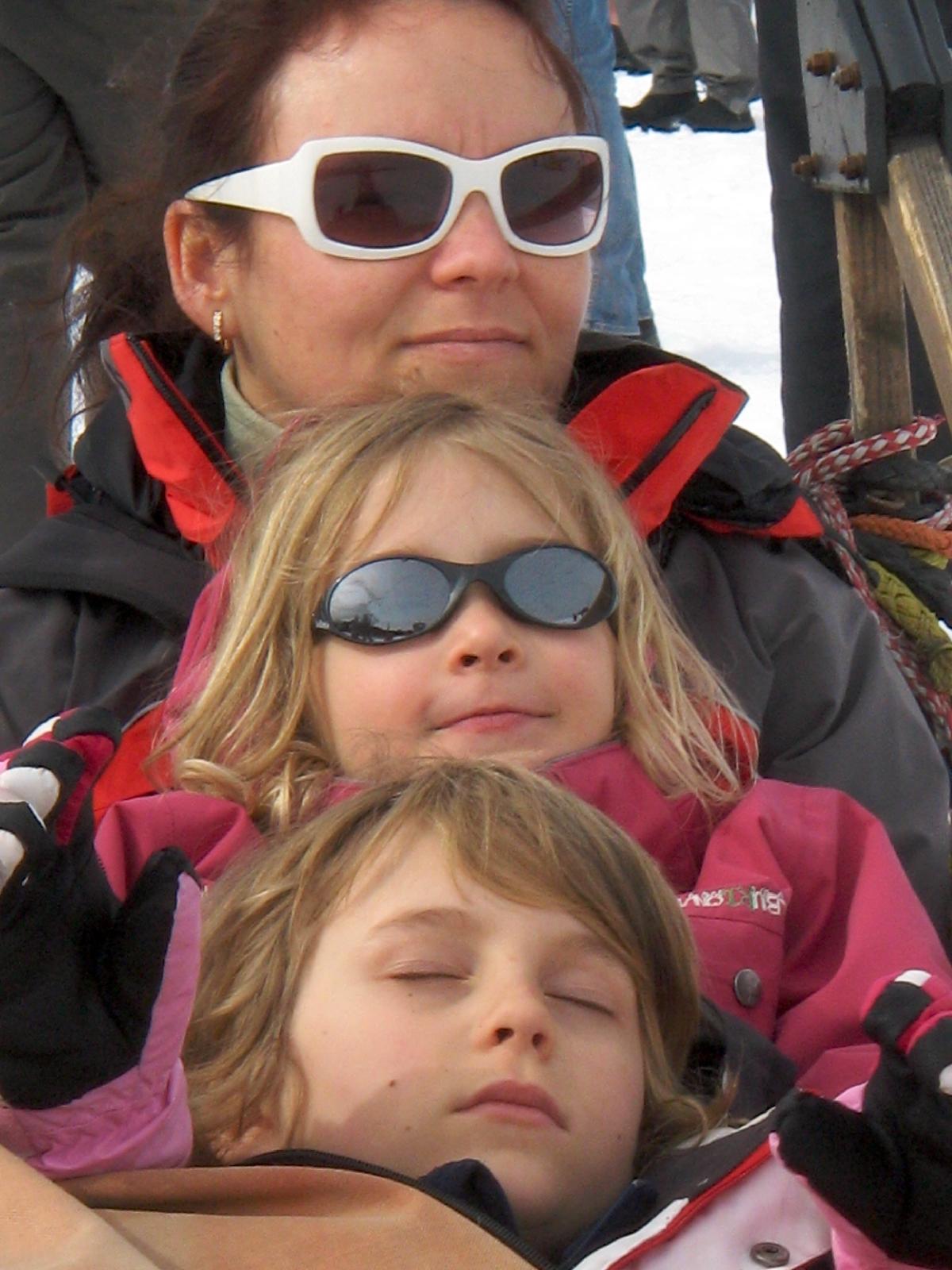}	
\includegraphics[height=50px]{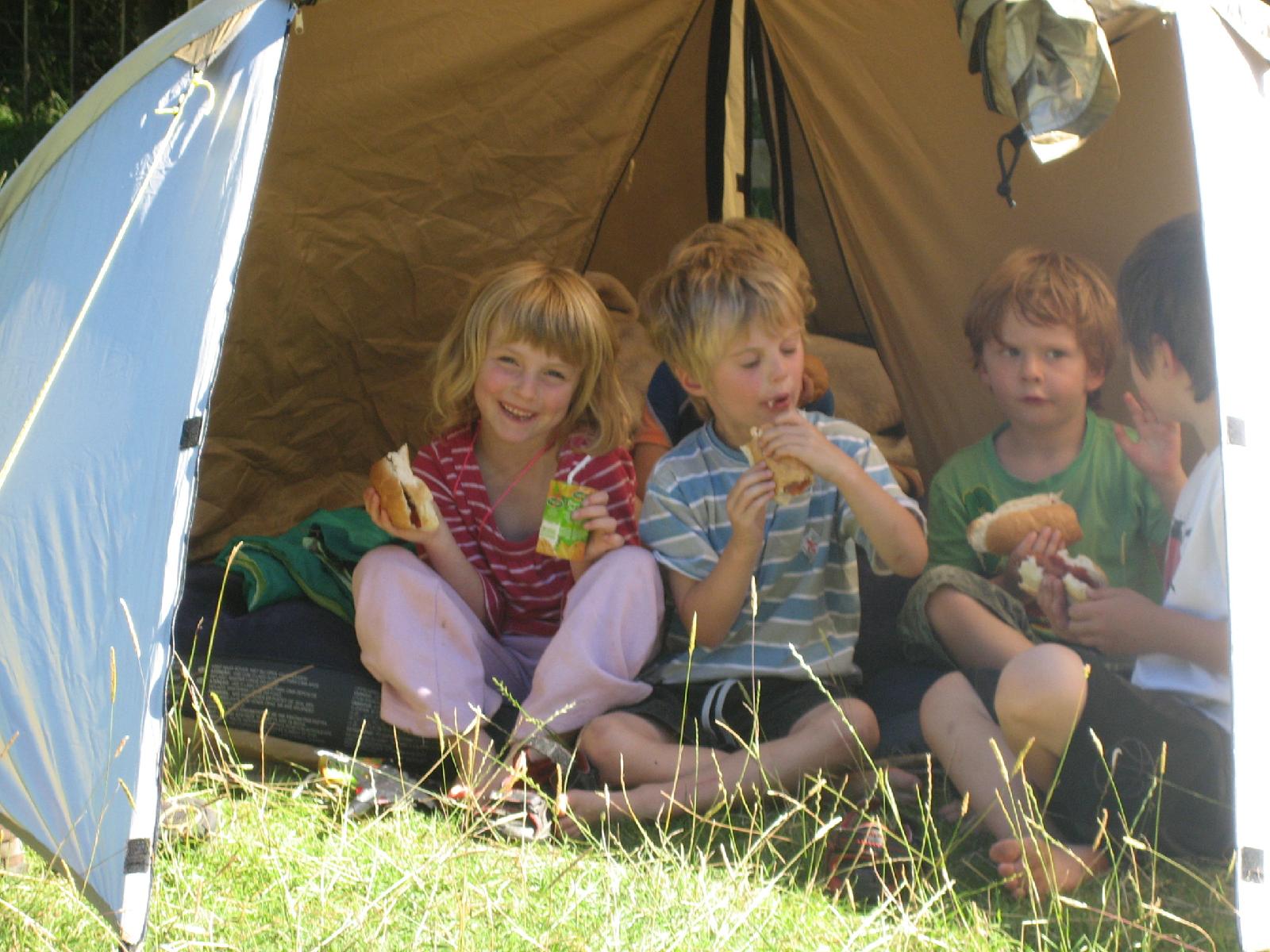}	
\includegraphics[height=50px]{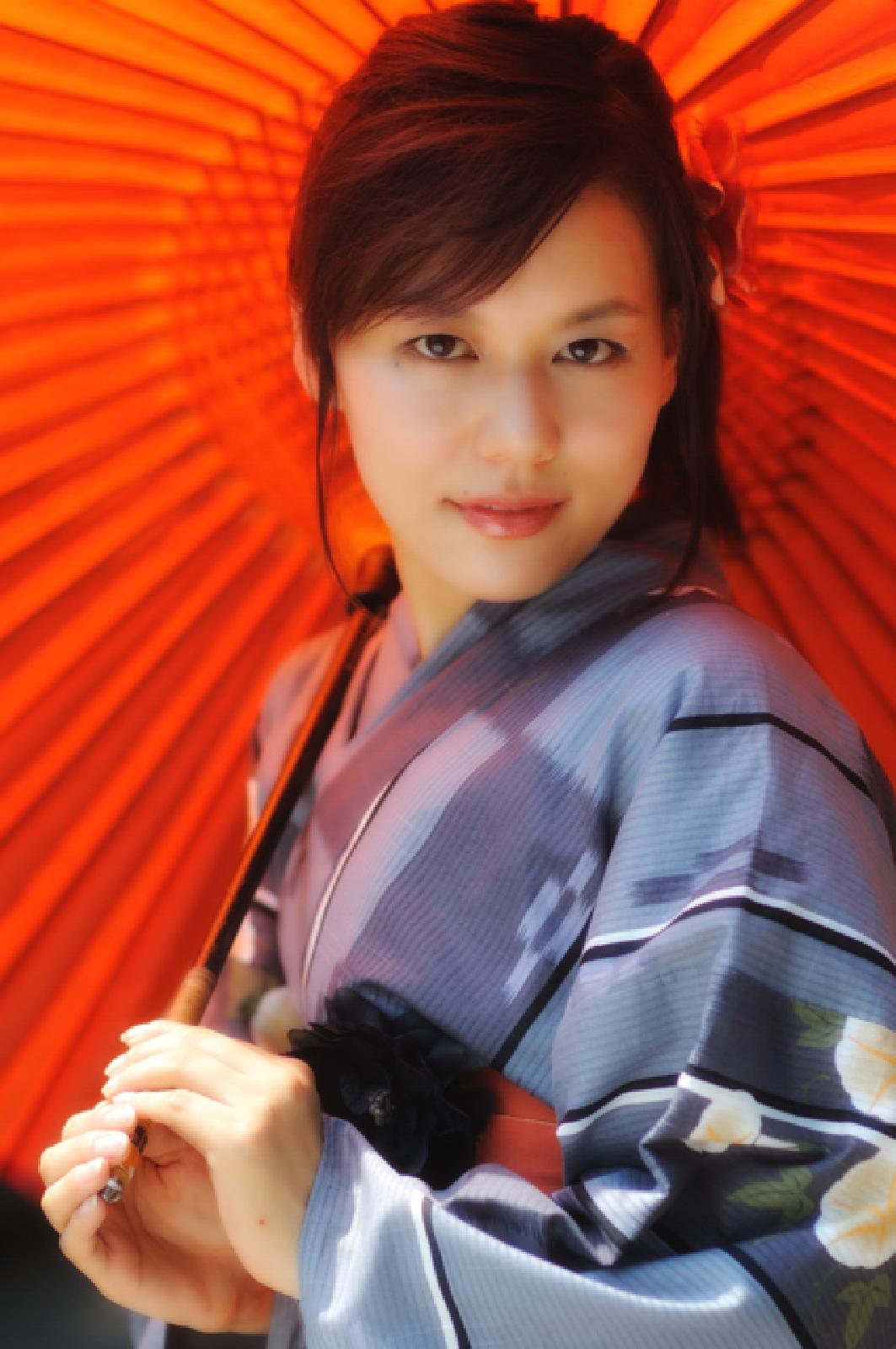}	
\includegraphics[height=50px]{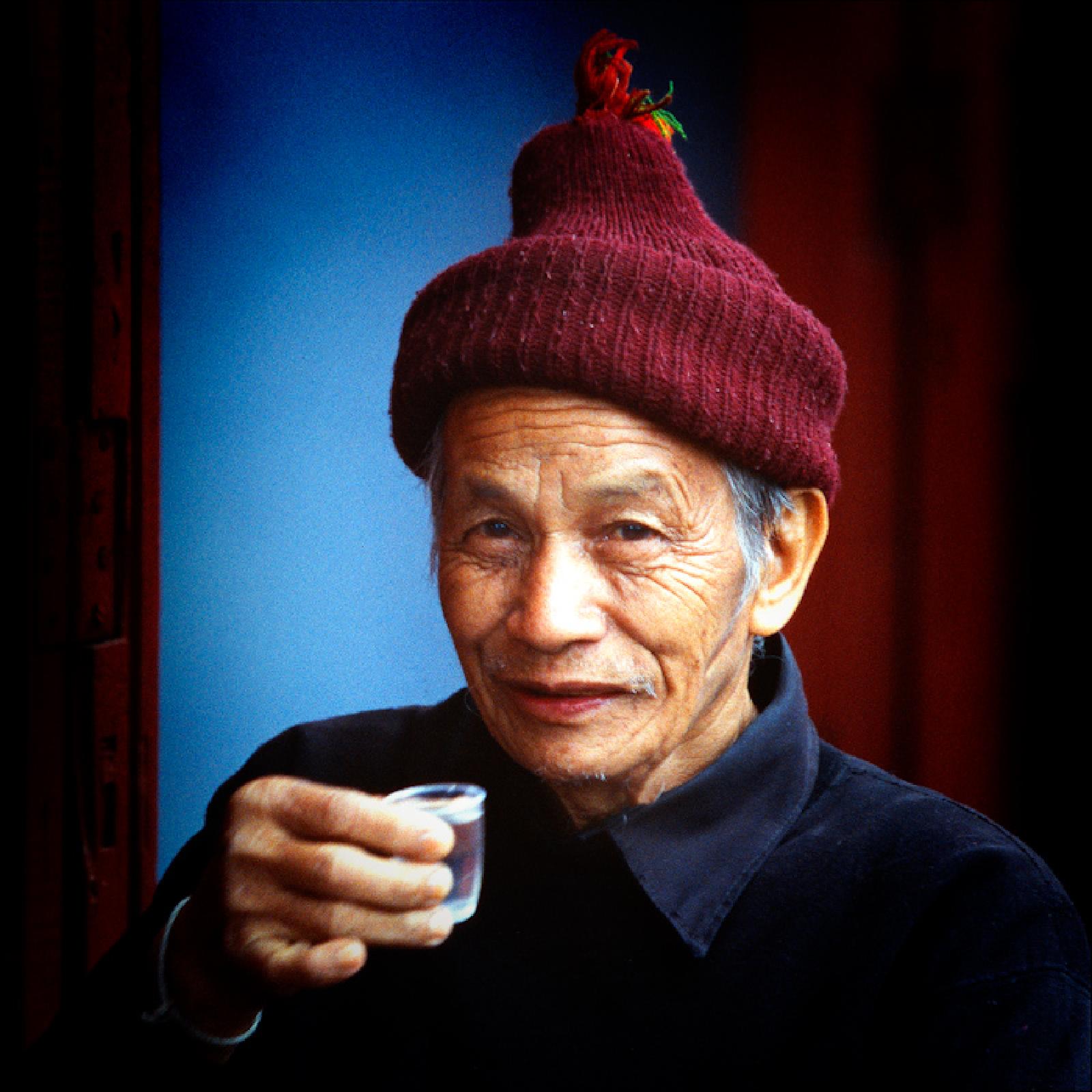}	
\begin{tikzpicture}
\node[draw=none,fill=none] (B) at (210:4.6) {0};
\node[draw=none,fill=none] (C) at (330:4.6) {10};
\draw[latex'-latex',double] (B) -- (C);
\end{tikzpicture}
\caption{Face images from the Flickr database.}
\label{flickr-samples}
\end{subfigure}
\caption{Examples of face images from the considered databases.}
\label{db-samples}
\end{figure}
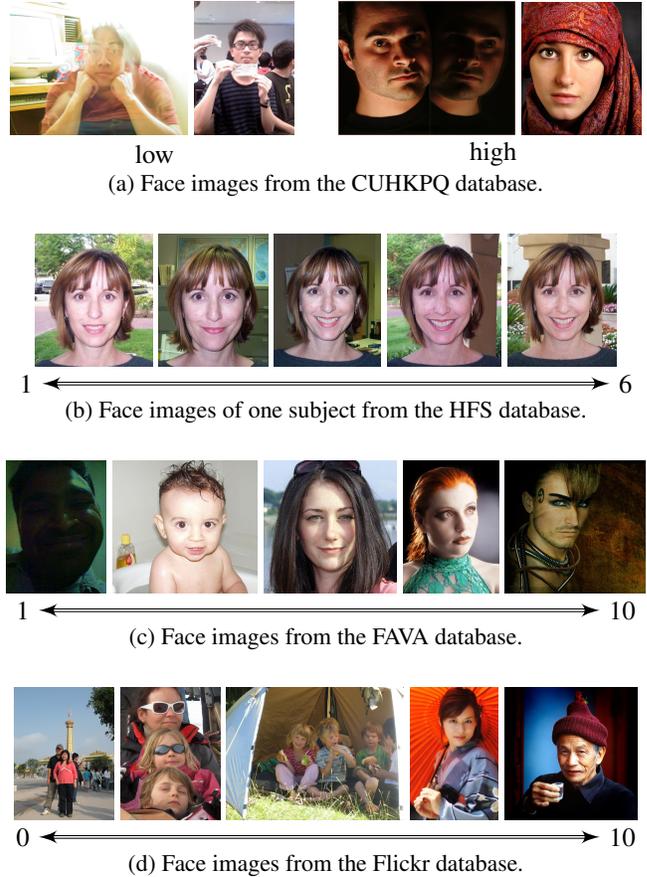
\begin{table}[ht]
\centering
\caption{GCR (\%) of the aesthetic quality categorization for each database by extracting perceptual features from the whole image.}
\label{result-gcr-wctx}
\resizebox{!}{5.1em}{
\begin{tabular}{cccccccc}
\hline
\multirow{2}{*}{IQ} & \multirow{2}{*}{IA} & \multirow{2}{*}{FA} & \multirow{2}{*}{\#features}	& \multirow{2}{*}{GA} & \multicolumn{3}{c}{GCR (\%)} \\
               		& 					  & 					&              	            	&                     & CUHKPQ   & FAVA   & Flickr   \\	\hline
    \checkmark      &  					  &                     &   		4,096				& 					  &	  93.2   &  63.6  &   64.3   \\
                    & 	\checkmark        &    			        &   		4,096				& 					  &   97.2   &  67.4  &   71.6   \\
                    &					  &		\checkmark		&   		2,048				& 					  &	  97.0	 &  70.0  &	  66.2	 \\
  \checkmark        &  					  &     \checkmark      &   		6,144				& 					  &   97.2   &  70.0  &   67.6   \\
   \checkmark       &  	\checkmark		  &                     &   		8,192				& 					  &   97.4   &  63.0  &   73.6   \\
       			    &  	\checkmark		  &     \checkmark      &   		6,144				& 					  &   98.2   &  71.2  &   73.6   \\
    \checkmark      & 	\checkmark 		  &     \checkmark      &   	   10,240				& 					  &   98.2   &  71.2  &   74.0   \\
    \checkmark      & 	\checkmark 		  &     \checkmark      &   8,300			&	  \checkmark	  &   97.5   & 	70.7  &   73.9   \\	\hline
\end{tabular}}
\end{table}
\begin{table}[ht]
\centering
\caption{LCC of the aesthetic quality prediction for each database by extracting perceptual features from the whole image.}
\label{result-lcc-wctx}
\resizebox{!}{5.1em}{
\begin{tabular}{ccccccc}
\hline
\multirow{2}{*}{IQ} & \multirow{2}{*}{IA} & \multirow{2}{*}{FA} & \multirow{2}{*}{\#features}	& \multirow{2}{*}{GA}	& \multicolumn{2}{c}{LCC} \\
               	 	&                     &                     &								&						&  FAVA   & Flickr  \\ \hline
   \checkmark       &  					  &                     &			4,096			    &   					&  0.38   &   0.36  \\
                    & 	\checkmark        &       			    &     		4,096				&						&  0.51   &   0.57  \\
                    &					  &		\checkmark		&	 		2,048				&						&  0.55   &	 0.48	\\
   \checkmark       &  					  &     \checkmark      &  			6,144				&						&  0.57   &   0.51  \\
   \checkmark       &  	\checkmark		  &                     &			8,192				&					    &  0.36   &   0.56  \\
       			    &  	\checkmark		  &     \checkmark      &  			6,144				&						&  0.62  &   0.62   \\
   \checkmark       & 	\checkmark 		  &     \checkmark      &  			10,240				&						&  0.61   &   0.61  \\
   \checkmark       & 	\checkmark 		  &     \checkmark      &  			10,229				&		\checkmark		&  0.62   &  0.61   \\ \hline
\end{tabular}}
\end{table}
\subsection{Experimental results}
In this section, experimental setup and results are detailed. Binary aesthetic classification and aesthetic score regression are performed for each dataset previously presented. For classification, datasets are separated in two equally distributed groups (except CUHKPQ which is already separated by labels), containing respectively the images with the lowest and highest aesthetic scores. For experiments based on the use of feature concatenation and SVM, we employ a linear SVM for binary classification while a linear Support Vector Regressor machine (SVR) is used for continuous aesthetic score prediction. We report the performance obtained by considering a single feature vector at time and then by all of their possible combinations. In the experiments involving the use of GA, all the feature vectors are linearly concatenated. For both classification and regression, the GA is trained with a population of 100 individuals initialized by using parameters (weights and bias) and their perturbed versions of the SVM previously learned for aesthetic prediction. The learning parameters are empirically setup differently for classification and regression. More precisely, for classification the number of generations is 200, the probability of crossover is 80\%, and the elitism is 7\%. For regression, the number of generation is 250, the crossover probability is 85\%, and finally the elitism is 10\%.

In order to evaluate how the context (background) influences the aesthetic judgement of images with faces, we perform two sets of experiments. In the first set, perceptual features are extracted from the whole image as previously described, while in the second set these features are extracted considering only the face region.

\textbf{Experiments considering the whole image.} Results for binary aesthetic classification are reported in Table \ref{result-gcr-wctx}. The combination of all the considered features achieved the best results for all the databases and performance results by the GA are very close but using a smaller set of features. Performance results for continuous aesthetic score are in Table \ref{result-lcc-wctx}. The best correlation is achieved for both FAVA and Flickr by fusing image aesthetics and facial attributes features.

\textbf{Experiments considering only face region.} Results for binary aesthetic classification are reported in Table \ref{result-gcr-woctx}. The performance for the FAVA dataset is higher than the one obtained by extracting features from the whole image. The reason might be that many images contain a small portion of background. Performance results (in Table \ref{result-lcc-woctx}) for continuous aesthetic score confirm that the fusion of all the features is optimal and that the GA-based solution obtains comparable results by using a smaller amount of features.
\begin{table}[ht]
\centering
\caption{GCR (\%) of the aesthetic quality categorization for each database by extracting perceptual features from face region.}
\label{result-gcr-woctx}
\resizebox{!}{5.1em}{
\begin{tabular}{ccccccccc}
\hline
\multirow{2}{*}{IQ} & \multirow{2}{*}{IA} & \multirow{2}{*}{FA} & \multirow{2}{*}{\#features}	& \multirow{2}{*}{GA} & \multicolumn{4}{c}{GCR (\%)} \\
               		& 					  & 					&              	            	&                     & CUHKPQ   &  HFS   &  FAVA   & Flickr  \\ \hline
    \checkmark      &  					  &                     &   		4,096				& 					  &	  92.0   &  72.4  &   63.3  &  59.1	  \\
                    & 	\checkmark        &    			        &   		4,096				& 					  &   95.0   &  73.8  &   66.5  &  64.5	  \\
                    &					  &		\checkmark		&   		2,048				& 					  &	  97.0	 &  71.0  &	  70.0	&  66.2   \\
  \checkmark        &  					  &    \checkmark       &   		6,144				& 					  &   97.0   &  76.8  &   70.8  &  67.2   \\
   \checkmark       &  	\checkmark		  &                     &   		8,192				& 					  &   95.4   &  75.1  &   66.3  &  65.0   \\
       			    &  	\checkmark		  &     \checkmark      &   		6,144				& 					  &   97.1   &  78.0  &   71.7  &  65.4	  \\
    \checkmark      & 	\checkmark 		  &     \checkmark      &   	   10,240				& 					  &   97.0   &  79.0  &   71.8  &  65.6   \\
    \checkmark      & 	\checkmark 		  &     \checkmark      &   8,283			&	  \checkmark	  &   96.1   & 	79.0  &   71.1  &	66.5 \\	\hline
\end{tabular}}
\end{table}
\begin{table}[ht]
\centering
\caption{LCC of the aesthetic quality prediction for each database by extracting perceptual features from face region.}
\label{result-lcc-woctx}
\resizebox{!}{5.1em}{
\begin{tabular}{cccccccc}
\hline
\multirow{2}{*}{IQ} & \multirow{2}{*}{IA} & \multirow{2}{*}{FA} & \multirow{2}{*}{\#features}	& \multirow{2}{*}{GA}	& \multicolumn{3}{c}{LCC} \\
               	 	&                     &                     &								&						&   HFS	  &   FAVA   & Flickr  \\ \hline
   \checkmark       &  					  &                     &			4,096			    &   					&  0.59   &   0.39	 &	0.32   \\
                    & 	\checkmark        &       			    &     		4,096				&						&  0.66   &   0.50   &  0.48   \\
                    &					  &		\checkmark		&	 		2,048				&						&  0.67   &	  0.55	 &  0.48   \\
   \checkmark       &  					  &     \checkmark      &  			6,144				&						&  0.71   &   0.56   &	0.49   \\
   \checkmark       &  	\checkmark		  &                     &			8,192				&					    &  0.68   &   0.51	 &  0.47   \\
       			    &  	\checkmark		  &     \checkmark      &  			6,144				&						&  0.74   &   0.62   &  0.51   \\
   \checkmark       & 	\checkmark 		  &     \checkmark      &  			10,240				&						&  0.74   &   0.61   &  0.51   \\
   \checkmark       & 	\checkmark 		  &     \checkmark      &  			10,087		&		\checkmark		&  0.76	  &   0.61	 &    0.51	   \\ \hline
\end{tabular}}
\end{table}

Table \ref{face-aesth-results} shows the comparison with state-of-the-art methods. We report results for the best solution on all the datasets corresponding to the combination of all the considered features extracted from the whole image. For all datasets, on average we improve GCR by more than 3\% with respect to the previous methods for binary aesthetic classification. The improvement in terms of LCC is more than 8\% on average.
\begin{table}
\centering
\caption{Comparison with state-of-the-art methods for both aesthetic categorization and score prediction for all the considered databases.}
\label{face-aesth-results}
\resizebox{\columnwidth}{!}{
\begin{tabular}{llllllll}
\hline
\multirow{2}{*}{Methods} & CUHKPQ & \multicolumn{2}{c}{HFS} & \multicolumn{2}{c}{FAVA} & \multicolumn{2}{c}{Flickr} \\
						 				&  GCR (\%)	&  GCR (\%)	&  	LCC		&  GCR (\%)	&  	LCC		&  GCR (\%)	&	LCC		\\	\hline
Lienhard \cite{lienhard2015predict}		&	94.8	& 	79.3	& 	0.73	&	67.1	&	0.51	&	69.3	&	0.49	\\
Kairanbay \cite{kairanbay2016aesthetic}	&	-		& 			& 			&	65.3	&	-		&	-		&	-		\\
Proposed							 	&	98.2	&  	79.0*	& 	0.76*	&	71.2	&	0.61	&	74.0	&	0.61	\\	\hline
\multicolumn{8}{l}{*These results are obtained by extracting perceptual features from face region.}
\end{tabular}}
\end{table}
\section{Conclusions}
\label{conclusions}
In this work, we propose a framework for the automatic estimation of the aesthetic quality of images containing faces. This work extends our generic-content aesthetic assessment framework specializing it for photo containing faces. We use three different CNNs to encode global image aesthetics, perceptual quality and facial attributes. A novel learning procedure based on genetic algorithms is then applied for the combination of CNNs features and image aesthetic prediction. We evaluate the proposed algorithm in both binary and continuous aesthetic score prediction tasks on four benchmark datasets achieving state-of-the-art performances.
\section{Acknowledgments}
We gratefully acknowledge the support of NVIDIA Corporation with the donation of the Titan X Pascal GPU used for this research.
%
\bibliographystyle{IEEEbib}
\bibliography{refs.bib}

\end{document}